\documentclass[sigconf]{acmart}
\renewcommand\footnotetextcopyrightpermission[1]{} 
\AtBeginDocument{%
  }

\setcopyright{acmlicensed}
\copyrightyear{2018}
\acmYear{2018}
\acmDOI{XXXXXXX.XXXXXXX}

\newcommand{\ourmodel}{HieraVid}
\usepackage{booktabs}
\usepackage{multirow}
\usepackage{array}
\usepackage{multicol}
\usepackage{dsfont}
\usepackage{dblfloatfix}
\usepackage{wrapfig}

\acmSubmissionID{6396}

\begin{document}
\pagestyle{plain}
\title{HieraVid: Hierarchical Token Pruning for Fast Video Large Language Models}

\author{Yansong Guo}
\affiliation{%
  \institution{Key Laboratory of Multimedia Trusted Perception and Efficient Computing,}
  \institution{Ministry of Education of China Xiamen University}
  \city{Xiamen}
  \state{Fujian}
  \country{China}}
\email{guoyansong@stu.xmu.edu.cn}

\author{Chaoyang Zhu}
\affiliation{%
  \institution{Key Laboratory of Multimedia Trusted Perception and Efficient Computing,}
  \institution{Ministry of Education of China Xiamen University}
  \city{Xiamen}
  \state{Fujian}
  \country{China}}
\email{sean.zhuh@gmail.com}

\author{Jiayi Ji}
\affiliation{%
  \institution{Key Laboratory of Multimedia Trusted Perception and Efficient Computing,}
  \institution{Ministry of Education of China Xiamen University}
  \city{Xiamen}
  \state{Fujian}
  \country{China}}
\email{jjyxmu@gmail.com}

\author{Jianghang Lin}
\affiliation{%
  \institution{Key Laboratory of Multimedia Trusted Perception and Efficient Computing,}
  \institution{Ministry of Education of China Xiamen University}
  \city{Xiamen}
  \state{Fujian}
  \country{China}}
\email{hunterjlin007@gmail.com}

\author{Liujuan Cao\footnotemark[2]}
\affiliation{%
  \institution{Key Laboratory of Multimedia Trusted Perception and Efficient Computing,}
  \institution{Ministry of Education of China Xiamen University}
  \city{Xiamen}
  \state{Fujian}
  \country{China}}
\email{caoliujuan@xmu.edu.cn}


\begin{abstract}
Video Large Language Models (VideoLLMs) have demonstrated impressive capabilities in video understanding, yet the massive number of input video tokens incurs a significant computational burden for deployment. Existing methods mainly prune video tokens at input level while neglecting the inherent information structure embedded in videos and large language models (LLMs). To address this, we propose HieraVid, a hierarchical pruning framework that progressively and dynamically reduces visual redundancy. Based on two observations that videos possess the segment-frame structure and LLMs internally propagate multi-modal information unidirectionally, we decompose pruning into three levels: 1) segment-level, where video tokens are first temporally segmented and spatially merged; 2) frame-level, where similar frames within the same segment are jointly pruned to preserve diversity; 3)  layer-level, redundancy gradually shrinks as LLM layer increases \textit{w/o} compromising performance. We conduct extensive experiments on four widely used video understanding benchmarks to comprehensively evaluate the effectiveness of \ourmodel. Remarkably, with only 30\% of tokens retained, \ourmodel{} achieves new state-of-the-art performance, while maintaining over 98\% and 99\% of the performance of LLaVA-Video-7B and LLaVA-OneVision-7B, respectively.
\end{abstract}

\maketitle
\renewcommand{\thefootnote}{\fnsymbol{footnote}}
\footnotetext[2]{Corresponding Author.}

\section{Introduction}
\label{sec:intro}

In recent years, multi-modal large language models (MLLMs)~\cite{qwen25vl,internvl35,video-llava,video-llama} have significantly advanced multi-modal understanding by jointly modeling visual and textual signals within a unified framework. These architectures typically adopt a modular design, where a pretrained visual encoder~\cite{clip,siglip2} extracts dense frame-level representations, which are subsequently aligned with the token space of an LLM~\cite{gpt3,llama,gpt4,qwen3} through a lightweight projector.
However, in the realm of video understanding, the temporal dimension introduces a substantial increase in token volume. Specifically, densely sampled frames produce a large number of visual tokens, among which a considerable portion corresponds to temporally redundant or spatially uninformative regions. These tokens are nevertheless uniformly processed by the LLM, implicitly assuming equal semantic importance across all spatiotemporal positions. This assumption leads to repeated encoding of temporally static and spatially unimportant video tokens within LLMs, introducing pronounced visual redundancy and inefficient utilization of computational resources.

Given the quadratic complexity of the prevalent self-attention mechanism~\cite{sa} in LLMs, the excessive token length further induces a super-linear increase in memory consumption and computational cost. As the number of tokens scales, attention computation becomes the primary bottleneck, significantly prolonging inference latency. This issue is particularly critical in long-video or real-time scenarios, where efficient processing is essential. These challenges highlight the necessity of redundancy-aware token modeling, aiming to selectively preserve informative tokens while suppressing irrelevant ones, thereby achieving a more effective trade-off optimization between computational efficiency and model performance without compromising critical semantic information.

\begin{figure}
    \centering
    \includegraphics[width=\linewidth]{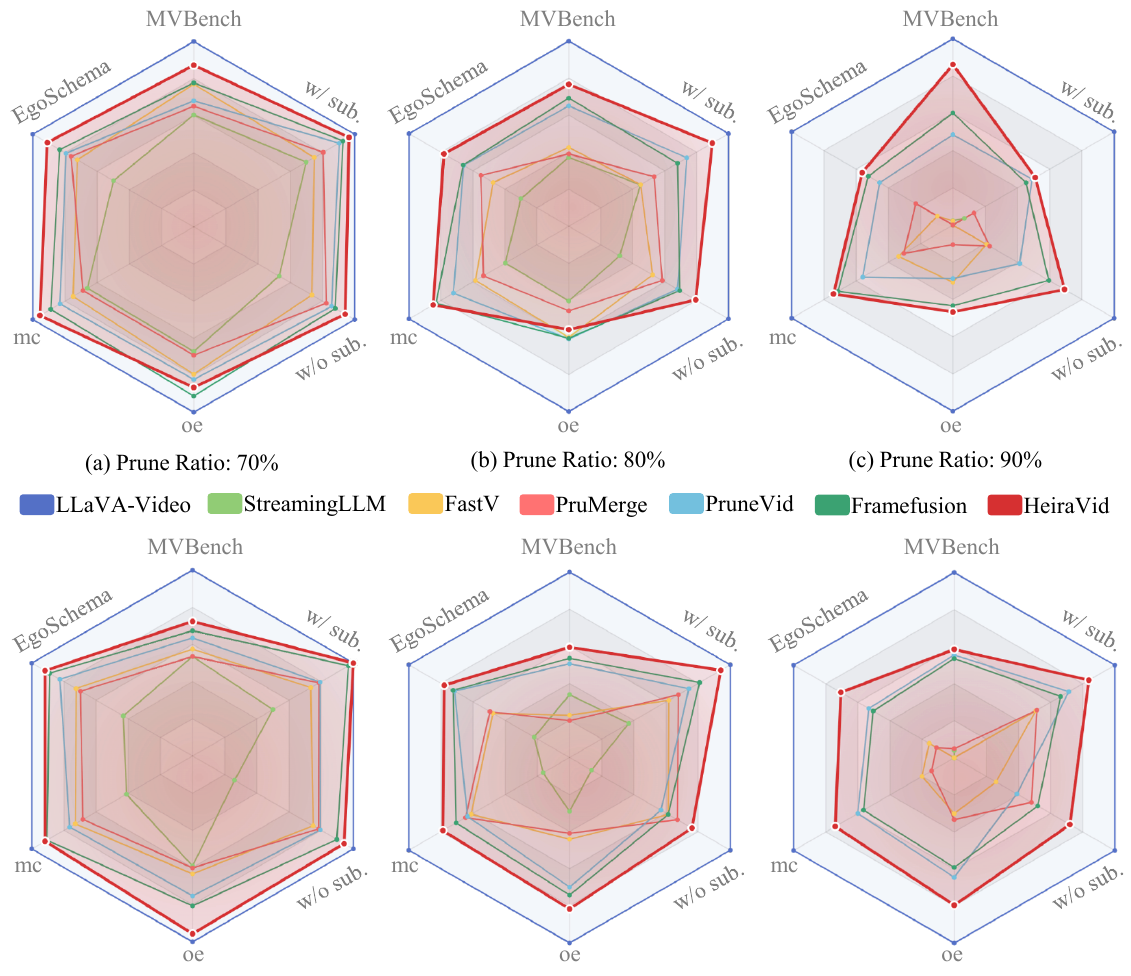}
    \caption{Comparison of existing VideoLLM pruning methods and our approach under different pruning ratios on LLaVA-Video (top) and LLaVA-OneVision (bottom).
    ``mc'' and ``oe'' represent the multiple choice and open ended settings in NExT-QA. ``w/o sub.'' and ``w/ sub.'' are the VideoMME settings without and with subtitles. Our model HieraVid is superior across all benchmarks with diverse durations, complexities, and pruning ratios.
}
    \label{fig:placeholder}
\end{figure}

Existing approaches of pruning MLLMs can be categorized in two ways: 1) where the pruning is applied, either before or within the LLM, and 2) what metric is used to assess the importance of visual tokens, typically attention scores or cosine similarities. Methods~\cite{prumerge,visionzip} that prune before LLMs effectively reduce spatial redundancy within individual frames, but they often ignore the temporal redundancy~\cite{fu2024framefusion,prunevid,mmg-vid,holitom} across frames, which is a unique characteristic to VideoLLMs. In contrast, pruning inside LLMs~\cite{fastv,pyramiddrop,sparsevlm} mainly removes redundant visual tokens after a chosen LLM layer, yet it fails to consider the internal multi-modal information flow~\cite{mminfoflow,labelwords} within LLMs. Moreover, though attention score and similarity serve as two useful proxy metrics for estimating the performance impact after removing a given visual token, they~\cite{dart,divprune,zhang2025beyond} usually require additional token merging or recycling steps to mitigate the information loss caused by the insufficient representational capacity of remaining tokens.

To overcome these limitations, we propose a hierarchical token pruning framework with three levels for VideoLLMs. Our first observation is that videos naturally exhibit semantic structures with boundaries separating segments of similar frames. Hence we first partition videos into distinct segments before LLM processing, enabling structure-aware token management. At frame-level, to prevent information loss, we introduce a segment-based determinantal point process (DPP)~\cite{dpp} that maximizes the diversity of frames within each segment. In this way, we ensure a more balanced visual token selection that collectively represents the semantics of each segment. Our second insight is that within the LLM, multi-modal information transfers in an upward-right direction~\cite{mminfoflow} through layers via causal self-attention. This causes visual tokens to become less discriminant~\cite{fu2024framefusion} toward the deeper layers of LLMs. In light of this, at layer-level, we design a progressively reduction strategy to optimize the trade-off between performance and redundancy.

As shown in Figure~\ref{fig:placeholder}, empirical evaluations validate the effectiveness of our method. 
On LLaVA-OneVision-7B model, \ourmodel~prunes 30\% of visual tokens while retaining 99.3\% of the original performance. Even with 90\% token pruning, it preserves 94.8\% of the performance, with computational costs reduced to just 9.3\% of the original FLOPs. These results demonstrate the robustness and effectiveness of our hierarchical pruning framework.

In summary, our key contributions are as follows:

\begin{itemize}
\item We leverage the inherent structural properties of videos and propose a novel merge-ratio guided segmentation method at segment-level, which partitions videos into semantically coherent segments. We further introduce a segment-based DPP pruning strategy at frame-level to select diverse and representative frames within each segment, reducing redundancy while preserving critical visual information.
\item Based on the observation of multi-modal information flow within LLMs, we further propose a multi-stage pruning strategy at layer-level to optimize the trade-off between final model performance and computational efficiency. 
\item Extensive experimental results demonstrate that our method consistently achieves state-of-the-art performance in a plug-and-play manner across various VideoLLMs and benchmarks, highlighting its strong generalization ability and practical effectiveness without additional training.
\end{itemize}

\begin{figure*}[ht]
  \centering
  \includegraphics[width=0.95\linewidth]{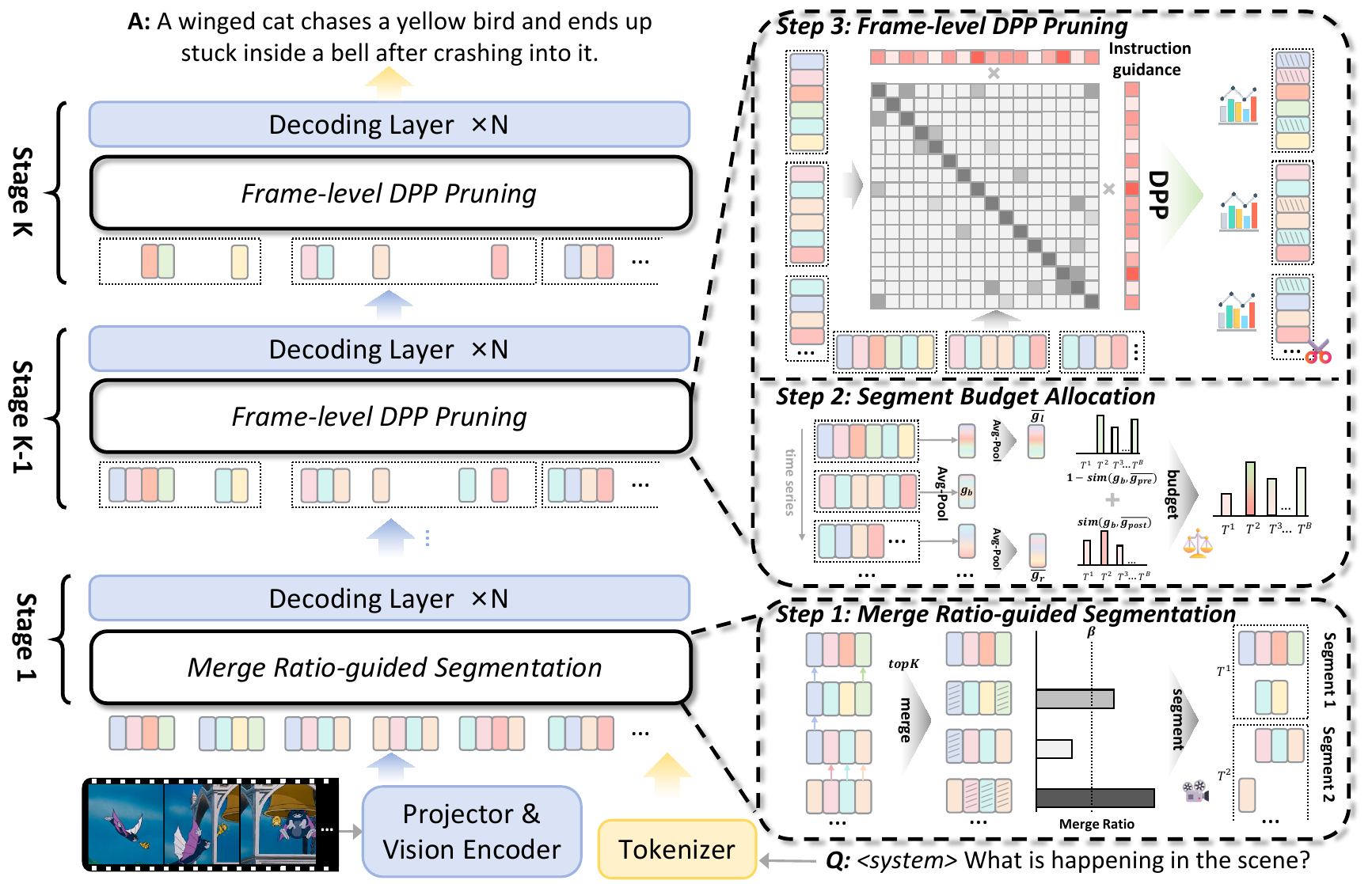}
  \vspace{-2.5mm}
  \caption{
  \textbf{Framework of~\ourmodel.}
  To balance visual information loss from shallow-layer pruning and inefficiency from deep-layer pruning, we apply layer-level multi-stage pruning to the LLM.
  \ourmodel~divides the pruning process into three stages.
  \textbf{(i) Merge Ratio-guided Segmentation:} At the LLM input layer, static tokens at corresponding positions across merged frames are temporally combined by merging similar tokens at their first occurrence. Furthermore, we partition frames into multiple segments based on the merge ratio between adjacent frames to ensure inter-segment diversity and intra-segment continuity (see Figure~\ref{fig:visualization_segment}).
  \textbf{(ii) Segment Budget Allocation} dynamically assigns pruning ratios to each segment to maximize visual diversity after pruning. 
  \textbf{(iii) Frame-level DPP Pruning} integrates the DPP kernel matrix with instruction features to balance visual diversity and instruction relevance in the pruned tokens.
  }
\label{fig:model}
\end{figure*}

\section{Related Work}
\label{sec:related}

\subsection{Video Large Language Models}
Recent advances in MLLMs have extended their understanding and reasoning capabilities from static image scene to dynamic video content, leading to the emergence of VideoLLMs~\cite{tao2025plug,ataallah2024minigpt4,song2024moviechat,wang2024qwen2,bai2025qwen2,fu2025video,jin2024video,zhang2023video,li2024llava,shu2025video,li2023videochat,xu2024pllava,li2024mvbench,li2024videochat}. They are typically composed of a visual encoder and an LLM, where sampled video frames are embedded into video tokens and fused with text for joint spatio-temporal reasoning~\cite{li2024llava,jin2024efficient,zhang2410video,yao2024minicpm,chen2024internvl,li2024llama,weng2024longvlm}. 
The long sequences of visual tokens produced by continuous video frames present a major obstacle to the practical scalability of video LLMs. Previous studies have explored various solutions to this issue. For instance, VideoLLaMA~\cite{video-llama} employs a Q-Former~\cite{li2023blip} to aggregate video tokens, and LLaVA-OneVision~\cite{li2024llava} applies pooling operations to reduce token redundancy.
LongVU~\cite{shen2024longvu} leverages cross-modal queries and inter-frame dependency modeling to adaptively identify and reduce redundant information across video frames, effectively enhancing temporal efficiency and representation compactness.
Video LLMs with training-time compression~\cite{song2024moviechat,xu2024pllava,shen2024longvu} aim to significantly reduce the number of video tokens, enabling longer video sequences. 
However, these approaches often introduce additional overhead, increasing inference time and memory consumption due to the extra complexity involved in the compression process.
This highlights the urgent need for efficient, training-free token compression strategies tailored for video LLMs, eliminating the dependence on costly fine-tuning and heavy hardware resources.
To this end, recent works have increasingly focused on training-free pruning methods that directly remove redundant tokens during inference, aiming to alleviate the reasoning bottleneck while maintaining a better trade-off between efficiency and performance.
\subsection{Visual Token Pruning}
Reducing redundant visual tokens has emerged as a critical approach for efficient VideoLLMs. Early strategies merge similar tokens either spatially or temporally~\cite{bolya2022token,shen2024tempme}, or prune unimportant tokens based on attention within visual encoders or LLMs~\cite{shang2025llava,yang2025visionzip,chen2024image}. 
An increasing number of works have explored training-free paradigm to improve inference efficiency~\cite{zhang2024sparsevlm,liu2024multi,pei2025greedyprune,liu2025video,wang2025dynamic,wen2025stop}. 
VisionZip~\cite{yang2025visionzip} selects important tokens using the attention from [CLS] token in the visual encoder. 
DivPrune~\cite{alvar2025divprune} and CDPruner~\cite{zhang2025beyond} go beyond attention or similarity-based proxy metrics, instead they focus on the diversity of retained visual tokens.
For pruning video tokens, PruneVid~\cite{prunevid} classifies tokens as static or dynamic and applies separate pruning strategies to each group.
FastV~\cite{fastv} detects unimportant visual tokens by the degree of cross-attention they received from instruction tokens within LLMs, which are dropped after an early LLM layer.
FrameFusion~\cite{fu2024framefusion} shows that token similarity becomes more concentrated in deeper LLM layers, hence it employs a two-stage framework of merging-then-pruning.
DyCoke~\cite{tao2025dycoke} first performs merging then applies a dynamic reduction of the KV cache. 
%
Inspired by PyramidDrop~\cite{xing2024pyramiddrop}, we also divide the LLM into multiple stages. However, we decompose pruning into three fine-grained levels including segment-level and frame-level, pushing the limit of trade-off between performance and efficiency.

\section{Methods}

We propose a hierarchical video pruning framework, HieraVid, which mines the information structure embedded in both videos and LLMs. Section~\ref{segment-level} first introduces how to partition videos into segments and allocate pruning budget for each segment. Then we detail how to preserve intra-segment diversity via segment-based determinantal point process in Section~\ref{frame-level}. Finally, we progressively reduce visual redundancy at layer-level leveraging multi-modal information flow in Section~\ref{layer-level}.

\begin{figure*}[t]
  \centering
  \includegraphics[width=1.0\linewidth]{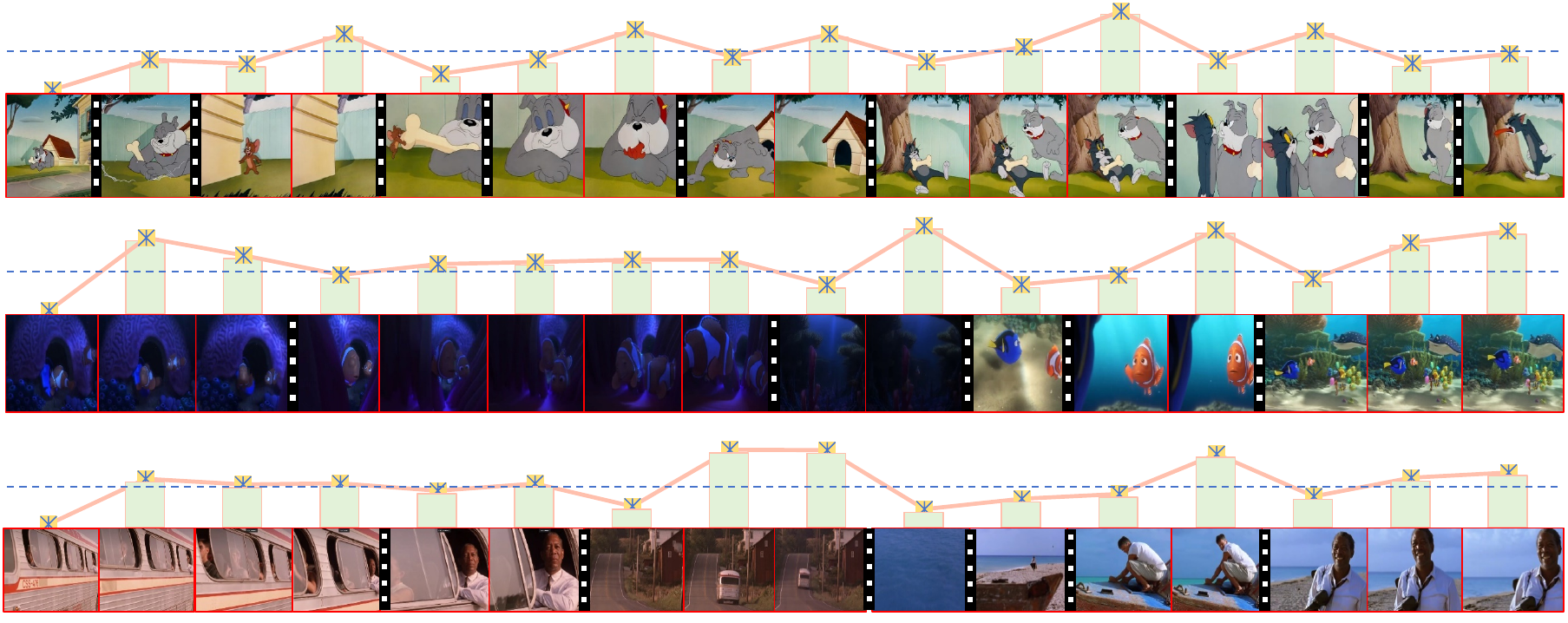}
  \caption{
  \textbf{Visualization of frame segment results by our \ourmodel.} The horizontal dashed line represents the segment threshold. Lower bars indicate lower similarity with preceding frames. Black borders mark the segmentation boundaries, where frames within each segment exhibit high similarity, while those across segments maintain distinctiveness.
}
\label{fig:visualization_segment}
\end{figure*}

\subsection{Segment-level Merging}
\label{segment-level}

\subsubsection{Merge Ratio-guided Segmentation}

Videos are inherently structured with segments composed of consecutive and similar frames. Given a video, we aim to produce $M$ segments $\{T_1, T_2, ..., T_M\}$, each segment $T_i$ contains variable frames $\{X_t\} \in \mathbb{R}^{N_m \times H \times W \times 3}$ where $N_m$ denotes the number of frames in segment $T_{i}$ and $t$ denotes the index of the $t$-th frame in the video sequence.

In contrast to existing coarse segmentation strategies~\cite{prunevid,mmg-vid} that rely on the global feature of each frame, we pursue a more fine-grained segmentation strategy based on local contextual relationships between frames. Each frame is first encoded by a visual encoder to yield visual tokens $F_t \in \mathbb{R}^{H' \times W' \times C}$. 
Let $S_t\in\mathbb{R}^{H' \times W'}$ denote the spatial similarity map between corresponding visual tokens in frame $t$ and frame $t-1$. For each location in the similarity map, $S_t(i,j)$ is computed as follows:
\begin{equation}
S_t(i,j) = \frac{F_t(i,j)F_{t-1}^{T}(i,j)}{\|F_t(i,j)\|_2\|F_{t-1}(i,j)\|_2},
\label{method:m1}
\end{equation}
where $T$ is transpose. We next define the overlap ratio $O_t$ between frame $t$ and frame $t-1$ as:
\begin{equation}
O_t = \frac{1}{H' * W'} \sum_{i=1}^{H'} \sum_{j=1}^{W'} \mathds{1}\big[(i,j,t) \in \text{Top-}K(S)\big],
\label{method:m2}
\end{equation}
%
where $\mathds{1}[\cdot]$ is an indicator function that evaluates to 1 if true otherwise 0. A boundary is thus created when the overlap ratio $O_t$ between frame $X_t$ and its previous frame falls below a threshold $\beta$. By repeating this process for each frame, we obtain $M$ segments.

After segmentation, we perform spatial merging across the whole video to eliminate static visual tokens, \textit{i.e.}, tokens with the same spatial position but not varied along temporal dimension.
Hence, a token $(i,j)$ is considered for merging in frame $t$ if it belongs to the Top-$K$ most similar tokens at the same spatial position across consecutive frames, where $K = R_{\text{merge}} \cdot N$ and $N$ is the total number of video tokens. For each spatial position, we compute the average of these $K$ selected tokens:
\begin{equation}
\hat{F}_{t_0}(i,j) = \frac{1}{K} \sum_{k=0}^{K-1} F_{t_0 + k}(i,j),
\label{method:m3}
\end{equation}
where $t_0$ is the first frame such that the following $K$ consecutive frames are among the Top-$K$ most similar tokens for position $(i,j)$, but not for the $t_0+K$-th frame, marking the end of this spatially redundant sequence. $\hat{F}_{t_0}(i,j)$ replaces the first token in this sequence, while all other tokens at the same location within the sequence are discarded. This ensures that spatially redundant tokens are merged according to the merge ratio $R_{\text{merge}}$.
%

%
\subsubsection{Segment Budget Allocation}
As shown in Figure~\ref{fig:visualization_segment}, after the Merge Ratio-guided Segmentation divides the video into multiple segments, each segment contains a varying number of frames and exhibits clear inter-segment diversity. Therefore, we compute a distinct pruning budget for each segment, assigning different pruning ratios accordingly. Such a design enables more fine-grained token allocation, ensuring that segments with richer or more diverse content retain more informative frames, while redundant segments are more aggressively pruned, leading to a balanced representation.

%
For all segments $\{T_1, T_2, ..., T_M\}$, we perform average pooling over all visual tokens within each segment $T_m$ to obtain its representation $g_m$.
Typically, earlier segments prioritize diversity while later segments prioritize representativeness. This observation naturally guides the allocation of the pruning budget for $T_m$:
\begin{equation}
b_m = 
\lambda * s(g_m,\overline{g_{r}})+(1-\lambda)*(1-s(g_m,\overline{g_{l}}))
\label{method:m4}
\end{equation}
where $\overline{g_{l}}=\sum_{i=1}^{m-1}g_i,\overline{g_{r}}=\sum_{i=m+1}^{M}g_i$, \textcolor{black}{$s$ denotes cosine similarity}, and $\lambda$ is a balancing hyperparameter. Hence, the prune rate for each segment $T_m$ \textcolor{black}{in stage $i$} is:
\begin{equation}
{R^{i,m}}=
{R^{i}_{prune}} + R_{var} * \frac{b_m - \text{mean}(b)}{\text{std}(b)},
\label{method:m6}
\end{equation}
here, $R^{i}_{prune}$ denotes the pruning ratio assigned to each stage $s_i$ as described in Section~\ref{layer-level}, $R_{\text{var}}$ represents the pruning deviation, and $b = \{ b_m \mid m = 1, 2, \dots, M \}$. The pruning ratio for each segment is dynamically adjusted according to its budget.

\begin{table*}[ht]
\centering
\small
\setlength{\tabcolsep}{6pt}
\renewcommand{\arraystretch}{1.15}
\resizebox{\linewidth}{!}{%
\begin{tabular}{c|cccccccc|ccc@{}}
\toprule
\multirow{2}{*}{\textbf{Method}} &
\multirow{2}{*}{\shortstack{\textbf{Prefilling}\\\textbf{FLOPs (T)}}} &
\multirow{2}{*}{\shortstack{\textbf{FLOPs}\\\textbf{Ratio}}} &
\multirow{2}{*}{\textbf{MVBench}} &
\multirow{2}{*}{\textbf{EgoSchema}} &
\multicolumn{2}{c}{\textbf{NExT-QA}} &
\multicolumn{2}{c}{\textbf{VideoMME}} &

\multicolumn{2}{c}{\textbf{Avg.}} & \\
\cmidrule(lr){10-11} \cmidrule(lr){6-7} \cmidrule(lr){8-9}
& & & &  &  \textbf{mc} & \textbf{oe\_test} &  \textbf{w/o sub.} & \textbf{w/ sub.} & \textbf{score} & \textbf{\%} \\
\midrule
LLaVA-Video &80.2&100\% 
&60.4 &59.4 &80.2 &34.6 &64.1 &71.4 &61.7 &100.0\% \\
\midrule
\multicolumn{12}{c}{\textbf{30\% relative token budget}} \\
\midrule
StreamingLLM (ICLR2024)~\cite{xiao2023efficient} &60.3&75.2\%
&53.8 &51.6 &76.1 &32.4 &55.5 &65.7 &55.9 &90.6\% \\  
FastV (ECCV2024)~\cite{fastv}&23.5&39.3\%
&56.6 &55.1 &77.2 &33.2 &59.3 &66.7 &58.0 &94.0\% \\ 
PruMerge (ICCV2025)~\cite{prumerge} &19.1&23.8\%
&54.6 &55.7 &76.4 &32.5 &60.9 &67.7 &58.0 &94.0\% \\  
PruneVid (ACL2025)~\cite{prunevid}&19.1&23.8\%
&55.1 &56.2 &78.2 &33.4 &61.5 &69.6 &59.0 &95.6\% \\  
Framefusion (ICCV2025)~\cite{fu2024framefusion}&19.1&23.8\%
&\underline{56.7} &\underline{56.8} &\underline{78.8} &\textbf{34.0} &\underline{61.9} &\underline{70.1} &\underline{59.7} &\underline{96.8\%} \\  
\textbf{HieraVid}&19.6&24.5\%
&\textbf{58.3} &\textbf{59.2} &\textbf{79.9} &\underline{33.7} &\textbf{62.3} &\textbf{70.8} &\textbf{60.7} &\textbf{98.4\%} \\
\midrule

\multicolumn{12}{c}{\textbf{20\% relative token budget}} \\
\midrule
StreamingLLM (ICLR2024)~\cite{xiao2023efficient}&56.7&70.7
&50.0 &48.5 &73.0 &30.6 &51.6 &61.0 &52.4 &85.0\% \\  
FastV (ECCV2024)~\cite{fastv}	&17.1&21.4\%
&50.9 &51.2 &75.2 &31.9 &55.4 &61.0 &54.4 &88.2\% \\  
PruMerge (ICCV2025)~\cite{prumerge}	&12.3&15.5\%
&50.3 &52.4 &74.6 &30.9 &56.5 &62.6 &54.6 &88.5\% \\  
PruneVid(ACL2025)~\cite{prunevid}	&12.3&15.5\%
&54.6 &\underline{55.1 }&77.0 &\textbf{32.3} &59.3 &66.2 &\underline{57.7} &\textbf{93.5\%} \\
Framefusion (ICCV2025)~\cite{fu2024framefusion}&12.3&15.5\%
&\underline{55.3} &54.1 &\underline{78.2} &31.9 &\underline{59.5} &\underline{66.2} &57.5 &93.3\% \\
\textbf{HieraVid}	&13.8&17.2\%
&\textbf{58.2} &\textbf{56.8} &\textbf{78.3} &\underline{32.2} &\textbf{61.8} &\textbf{66.3} &\textbf{58.9} &\textbf{95.5\%} \\

\midrule

\multicolumn{12}{c}{\textbf{10\% relative token budget}} \\
\midrule
StreamingLLM (ICLR2024)~\cite{xiao2023efficient}&52.1&65.0\%
&44.2 &43.8 &68.2 &27.9 &45.8 &53.9 &47.3 &76.7\% \\  
FastV (ECCV2024)~\cite{fastv}&11.21&14.0\%
&44.2 &45.4 &72.3 &29.9 &49.6 &52.5 &49.0 &79.4\% \\  
PruMerge (ICCV2025)~\cite{prumerge}&5.9&7.4\%
&43.8 &47.4 &71.9 &28.6 &50.0 &55.0 &49.4 &80.1\% \\  
PruneVid (ACL2025)~\cite{prunevid}&6.19&7.7\%
&\underline{53.9} &\underline{53.4} &75.2 &\underline{30.8} &\underline{57.7} &\underline{62.8} &\underline{55.6} &\underline{90.2\%} \\
Framefusion (ICCV2025)~\cite{fu2024framefusion} &6.1&7.6\%
&53.8 &53.0 &\underline{76.7} &30.4 &57.3 &61.7 &55.5 &89.9\% \\
\textbf{HieraVid} &6.85&8.5\%
&\textbf{58.1} &\textbf{53.7} &\textbf{77.1} &\textbf{31.3} &\textbf{58.8} &\textbf{63.2} &\textbf{57.0} &\textbf{92.4\%} \\
\bottomrule
\end{tabular}
}
\caption{\textbf{Comparison of state-of-the-art methods across video understanding benchmarks on LLaVA-Video-7B.} \textbf{Best} results are in bold, \underline{second best} underlined. “Avg.” indicates the mean performance across all benchmarks, where “score” represents the average score over all benchmarks, and “\%” denotes the ratio of this average score to that of the unpruned LLaVA-Video model.}
\label{tab:video_models}
\end{table*}

\subsection{Frame-level DPP Pruning}
\label{frame-level}
To enhance the representation of retained tokens, existing methods that estimate token importance using proxy metrics, such as attention scores or feature similarities, often require additional merging or recycling operations to compensate for the information discarded during pruning~\cite{divprune,dart}. Instead, we adopt a fundamentally different strategy by directly maximizing intra-segment token diversity via the DPP algorithm, which encourages the selection of a subset of tokens that are both representative and minimally redundant.

Formally, a DPP $\mathcal{P}$ defines a probability distribution over all subsets of a finite ground set $Z = {1, 2, \ldots, n}$. It is characterized by a positive semi-definite kernel matrix $L \in \mathbb{R}^{n \times n}$, 
such that for every subset $S \subseteq Z$, the probability of selecting a subset $S$ is:
\begin{equation}
\mathcal{P}(S) =
\frac{\det(L_S)}{\det(L + I)} \propto \det(L_S),
\label{method:m0}
\end{equation}
with $L_S$ denoting the principal submatrix of $L$ corresponding to $S$. This formulation naturally encourages diversity, as the determinant captures repulsion among similar items.
According to the DPP sampling process, the optimal subset $S^{*}$ is determined as follows:
\begin{equation}
S^* = \underset{S \subseteq Z, |S|=m}{\arg\max} \det(L_S).
\label{method:m00}
\end{equation}

To balance diversity and instruction-relevance, we incorporate instruction-related information into the DPP kernel matrix. Specifically, we use the last instruction token as a query to compute instruction relevance.
Given the visual embeddings $H_v \in \mathbb{R}^{n \times d}$ and the last instruction embedding $H_t \in \mathbb{R}^d$, we represent the relevance score of each visual token to the instruction as:
\begin{equation}
r = \text{min-max}\left(\text{softmax}\left ( \frac{H_tH_v^T}{\sqrt{d}} \right )\right),
\label{method:m7}
\end{equation}
where $\text{min-max}(\cdot)$ is min-max normalization to scale the relevance scores to $[0,1]$. The relevance scores are integrated into the visual-only kernel matrix $L = {H_v {H_v}^T}/{d}$ as:

\begin{equation}
\tilde{L}=\text{diag}\left (\tilde{r}  \right ) \times L \times \text{diag}\left (\tilde{r}  \right ).
\label{method:m9}
\end{equation}
Next, we split $\tilde{L}$ into $M$ sub-kernel matrices, each corresponding to a segmentation $T_m$:
\begin{equation}
\tilde{L} = \begin{bmatrix}
\tilde{L}_1 & {0} & \cdots & {0} \\
{0} & \tilde{L}_2 & \cdots & {0} \\
\vdots & \vdots & \ddots & \vdots \\
{0} & {0} & \cdots & \tilde{L}_M
\end{bmatrix},
\label{method:m10}
\end{equation}
where $\tilde{L}_m\in {R}^{n_m \times n_m}$, and $\sum_{m=1}^M n_m = |H_v|$.
For each sub-kernel $\tilde{L}_m$ in stage $i$, MAP inference selects an optimal subset of visual tokens of size $(k=R^{i,m} \cdot N_m)$.
Leveraging Cholesky decomposition, the pruning complexity is $\mathcal{O}(n_mk^2)$, and since $k < n_m$, this process completes efficiently within just milliseconds.

\subsection{Layer-level Balancing}
\label{layer-level}
{\color{black}
Inspired by previous works~\cite{xing2024pyramiddrop,zhang2025cross}, we observe that VideoLLMs process multi-modal information in a multi-stage manner when generating final predictions. In shallow layers, the model first transfers visual information to linguistic tokens. In middle layers, instruction-relevant visual tokens are then integrated to their corresponding instruction tokens. Finally in deep layers, the fused multi-modal representations are propagated to the final token position. Therefore, we structurally divide the LLM layers into $n$ stages with $S = \{s_i\}_{i=1}^{n}$.

Let $R_{prune}^i$ denote the pruning ratio for stage $s_i$ $(i \leq 2)$, $R_{merge}$ the merge ratio for stage $s_1$, and $F_i$ the corresponding decoder function comprising $L$ transformer decoder layers with pruning or merging operations.
In LLM forward pass, the hidden states of $s_i$ produce the visual and instruction tokens $H_v^i$ and $H_t^i$,formally:
\begin{equation}
H_v^i, H_t^i=F^s_i(H_v^{i-1}, H_t^{i-1}),
\label{method:m4}
\end{equation}
where $|H_v^i| = |H_v^0|(1-R_{merge}) \prod_{k=2}^{i-1} (1-R_{prune}^{k})$.
}

\begin{table*}[ht]
\centering
\small
\setlength{\tabcolsep}{6pt}
\renewcommand{\arraystretch}{1.15}
\resizebox{\linewidth}{!}{%
\begin{tabular}{c|cccccccc|ccc@{}}
\toprule
\multirow{2}{*}{\textbf{Method}} &
\multirow{2}{*}{\shortstack{\textbf{Prefilling}\\\textbf{FLOPs (T)}}} &
\multirow{2}{*}{\shortstack{\textbf{FLOPs}\\\textbf{Ratio}}} &
\multirow{2}{*}{\textbf{MVBench}} &
\multirow{2}{*}{\textbf{EgoSchema}} &
\multicolumn{2}{c}{\textbf{NExT-QA}} &
\multicolumn{2}{c}{\textbf{VideoMME}} &

\multicolumn{2}{c}{\textbf{Avg.}} & \\
\cmidrule(lr){10-11} \cmidrule(lr){6-7} \cmidrule(lr){8-9}
& & & &  &  \textbf{mc} & \textbf{oe\_test} &  \textbf{w/o sub.} & \textbf{w/ sub.} & \textbf{score} & \textbf{\%} \\
\midrule
LLaVA-OneVision &40.8 &100\% &58.4 &62.2 &79.4 &18.2 &58.5 &61.9 &56.4 &100.0\% \\
\midrule
\multicolumn{12}{c}{\textbf{30\% relative token budget}} \\
\midrule
StreamingLLM (ICLR2024)~\cite{xiao2023efficient} &30.6 &75\%
&54.4 &56.4 &73.5 &15.7 &52.0 &54.9 &51.1 &90.6\% \\ 
FastV (ECCV2024)~\cite{fastv}&12.7 &31.2\%
&57.0 &59.6 &76.2 &16.1 &56.3 &58.1 &53.9 &95.5\% \\
PruMerge (ICCV2025)~\cite{prumerge} &10.6 &25.9\%
&56.6 &59.3 &75.8 &15.8 &56.6 &58.8 &53.8 &95.3\% \\ 
PruneVid (ACL2025)~\cite{prunevid} &10.6 &25.9\%
&57.6 &60.7 &76.5 &17.2 &57.4 &59.9 &54.9 &97.2\% \\
Framefusion (ICCV2025)~\cite{fu2024framefusion}&10.6 &25.9\%
&\underline{58.0} &\underline{61.4} &\underline{77.7} &\textbf{18.2} &\underline{57.5} &\underline{61.1} &\underline{55.6} &\underline{98.6\%} \\
\textbf{HieraVid}&10.7 &26.3\%
&\textbf{58.1} &\textbf{62.6} &\textbf{78.0} &\underline{18.1} &\textbf{58.0} &\textbf{61.6} &\textbf{56.1} &\textbf{99.3\%} \\
\midrule

\multicolumn{12}{c}{\textbf{20\% relative token budget}} \\
\midrule
StreamingLLM (ICLR2024)~\cite{xiao2023efficient}&29.5&72.3\%
&50.6 &52.9 &71.5 &14.5 &49.5 &52.7 &48.6 &86.2\% \\  
FastV (ECCV2024)~\cite{fastv}	&9.3 &22.8\%
&52.9 &55.7 &74.6 &14.9 &53.7 &57.1 &51.5 &91.2\% \\ 
PruMerge (ICCV2025)~\cite{prumerge}	&6.9&16.9\%
&52.7 &55.9 &74.7 &15.3 &54.2 &56.5 &51.6 &91.4\% \\ 
PruneVid(ACL2025)~\cite{prunevid}	&6.9&16.9\%
&54.7 &59.8 &75.9 &16.8 &\underline{55.8} &58.4 &53.6 &94.9\% \\ 
Framefusion (ICCV2025)~\cite{fu2024framefusion} &6.9&16.9\%
&\underline{55.7} &\underline{60.7} &\underline{76.1} &\underline{17.0} &55.7 &\underline{60.1} &\underline{54.2} &\underline{96.1\%} \\  
\textbf{HieraVid}	&7.5  &18.5\%
&\textbf{56.5} &\textbf{61.5} &\textbf{77.4} &\textbf{17.9} &\textbf{57.1} &\textbf{60.7 }&\textbf{55.2} &\textbf{97.8\%} \\  

\midrule

\multicolumn{12}{c}{\textbf{10\% relative token budget}} \\
\midrule
StreamingLLM (ICLR2024)~\cite{xiao2023efficient}&26.9&65.9\%
&48.7 &50.6 &68.8 &13.5 &48.8 &51.2 &46.9 &83.2\% \\  
FastV (ECCV2024)~\cite{fastv}&6.037 &14.8\%
&50.3 &52.6 &71.2 &13.6 &51.7 &55.1 &49.1 &87.0\% \\ 
PruMerge (ICCV2025)~\cite{prumerge}&3.4& 8.3\%
&50.8 &53.1 &71.1 &13.5 &51.9 &54.7 &49.2 &87.2\% \\
PruneVid (ACL2025)~\cite{prunevid} &4.3 &10.6\%
&53.9 &\underline{58.3} &\underline{75.0} &\textbf{15.6} &\underline{53.7} &\underline{57.5} &\underline{52.3} &\underline{92.7\%} \\ 
Framefusion (ICCV2025)~\cite{fu2024framefusion} &3.4 &8.4\%
&\underline{54.0} &58.0 &74.8 &15.0 &53.5 &56.5 &52.0 &92.1\% \\
\textbf{HieraVid} &3.8 &9.3\%
&\textbf{55.6} &\textbf{58.7} &\textbf{76.0} &\underline{15.5} &\textbf{55.4} &\textbf{59.7} &\textbf{53.5} &\textbf{94.8\%} \\ 
\bottomrule
\end{tabular}
}
\caption{\textbf{Comparison of state-of-the-art methods across video understanding benchmarks on LLaVA-Onevision-7B.} \textbf{Best} results are in bold, \underline{second best} underlined. “Avg.” indicates the mean performance across all benchmarks, where “score” represents the average score over all benchmarks, and “\%” denotes the ratio of this average score to that of the unpruned LLaVA-Video model.}
\label{tab:video_models_ov}
\end{table*}

\section{Experiments}
\subsection{Experiment Setups}
\paragraph{Benchmarks.}
For consistency in comparison, we follow prior works~\cite{fu2024framefusion,mmg-vid,prunevid} to conduct evaluations on four widely used datasets, including NExT-QA~\cite{xiao2021next}, MVBench~\cite{li2024mvbench}, EgoSchema~\cite{mangalam2023egoschema}, and VideoMME~\cite{fu2025video}. These video benchmarks are carefully designed to encompass videos of diverse durations and complexities, offering a thorough and comprehensive evaluation of VideoLLMs, especially on their spatio-temporal reasoning ability.

\paragraph{Baselines.}
We compare the proposed HieraVid against the following approaches: 1) StreamingLLM~\cite{xiao2023efficient} that retains the KV cache of initial tokens to process infinite sequence lengths, eliminating the attention sink problem; 
2) FastV~\cite{fastv} measures token importance via attention scores and prunes visual tokens at an early layer of LLMs during prefilling stage;
3) PruMerge~\cite{prumerge} dynamically drops redundant visual tokens before LLMs and merges pruned ones to enhance the representational capacity of retained tokens;
4) PruneVid~\cite{prunevid} first performs spatial and temporal token clustering before LLMs, then selects key tokens based on text-to-video attention;
5) FrameFusion~\cite{fu2024framefusion} first merges tokens across frames according to similarity, then performs pruning similar to FastV~\cite{fastv}. These methods are all training-free and serve as strong baselines for fair comparison. Among them, FrameFusion~\cite{fu2024framefusion} is the state-of-the-art for pruning VideoLLMs currently.

\begin{table}[ht]
\centering
\small
\setlength{\tabcolsep}{6pt}
\renewcommand{\arraystretch}{1.15}
\resizebox{\linewidth}{!}{%
\begin{tabular}{c|cccc|ccc@{}}
\toprule
\multirow{2}{*}{\textbf{Method}} &
\multirow{2}{*}{\shortstack{\textbf{Prefilling}\\\textbf{FLOPs (T)}}} &
\multirow{2}{*}{\shortstack{\textbf{FLOPs}\\\textbf{Ratio}}} &
\multicolumn{2}{c}{\textbf{VideoMME}} &
\multicolumn{2}{c}{\textbf{Avg.}} & \\

\cmidrule(lr){4-5} \cmidrule(lr){6-7} &  &  & \textbf{w/o sub.} & \textbf{w/ sub.} & \textbf{score} & \textbf{\%} \\ 

\midrule
Vanilla	
&27.1 &100.0 &57.6 &61.0 &59.3 &100.0 \\

\midrule
PruneVid	
&7.3 &27.0 &53.8 &57.3 &55.6 &93.7 \\
Framefusion
&7.3 &27.0 &55.4 &59.7 &57.6 &97.0 \\
\ourmodel	
&7.4 &27.2 &\textbf{56.1} &\textbf{60.1} &\textbf{58.1} &\textbf{98.0} \\

\bottomrule
\end{tabular}
}
\caption{\textbf{Comparison of state-of-the-art methods on the VideoMME benchmark using Qwen2-VL} (Relative Token Budget: 30\%). Vanilla denotes the original Qwen2-VL before pruning. “Avg.” indicates the mean performance across all benchmarks, where “score” represents the average score over all benchmarks, and “\%” denotes the ratio of this average score to that of the unpruned LLaVA-Video model.}
\label{supp:qwen2vl}
\end{table}

\begin{table*}[ht]
\centering
\small
\setlength{\tabcolsep}{6pt}
\renewcommand{\arraystretch}{1.0}
\resizebox{0.8\linewidth}{!}{%
\begin{tabular}{c|cccccc|ccc@{}}
\toprule
\multirow{2}{*}{\textbf{Method}} &
\multirow{2}{*}{\textbf{MVBench}} &
\multirow{2}{*}{\textbf{EgoSchema}} &
\multicolumn{2}{c}{\textbf{NExT-QA}} &
\multicolumn{2}{c}{\textbf{VideoMME}} &

\multicolumn{2}{c}{\textbf{Avg.}} & \\
\cmidrule(lr){4-5} \cmidrule(lr){6-7} \cmidrule(lr){8-9}
& &  &  \textbf{mc} & \textbf{oe\_test} &  \textbf{w/o sub.} & \textbf{w/ sub.} & \textbf{score} & \textbf{\%} \\
\midrule
\ourmodel 
&58.3\phantom{$\downarrow$} &59.2\phantom{$\downarrow$} &79.9\phantom{$\downarrow$} &33.7\phantom{$\downarrow$} &62.3\phantom{$\downarrow$} &70.8\phantom{$\downarrow$} &60.7\phantom{$\downarrow$} &98.4\%\phantom{$\downarrow$} \\
\midrule
- Layer-level Balancing 
&57.8\textcolor{red}{$\downarrow$} &58.8\textcolor{red}{$\downarrow$} &78.8\textcolor{red}{$\downarrow$} &33.8\textcolor{green}{$\uparrow$} &61.1\textcolor{red}{$\downarrow$} &70.4\textcolor{red}{$\downarrow$} &60.1\textcolor{red}{$\downarrow$} &94.2\%\textcolor{red}{$\downarrow$} \\

- Merge Ratio-guided Segmentation
&57.8\textcolor{red}{$\downarrow$} &58.6\textcolor{red}{$\downarrow$} &78.5\textcolor{red}{$\downarrow$} &33.6\textcolor{red}{$\downarrow$} &59.0\textcolor{red}{$\downarrow$} &68.1\textcolor{red}{$\downarrow$} &59.3\textcolor{red}{$\downarrow$} &92.8\%\textcolor{red}{$\downarrow$} \\

- Frame-level DPP Pruning
&51.5\textcolor{red}{$\downarrow$} &50.7\textcolor{red}{$\downarrow$} &53.6\textcolor{red}{$\downarrow$} &25.5\textcolor{red}{$\downarrow$} &52.2\textcolor{red}{$\downarrow$} &56.1\textcolor{red}{$\downarrow$} &48.3\textcolor{red}{$\downarrow$} &78.2\%\textcolor{red}{$\downarrow$} \\

\bottomrule

\end{tabular}
}
\caption{\textbf{Ablation study on components of~
\ourmodel~}on LLaVA-Video (Retention Ratio: 30\%).}
\label{ablation:components}
\end{table*}

\paragraph{Inference Cost Evaluation.}
To evaluate the computational cost of different methods during inference, we focus on the transformer layers of the LLM. Since each layer primarily consists of multi-head attention (MHA) and feed-forward network (FFN) modules, we compute the FLOPs of these two components to approximate the actual computational cost. For a model with $L$ layers, let the number of visual tokens at the input of layer $i$ be $n_i$, the hidden size be $d$, and the FFN’s intermediate embedding size be $m$. The total FLOPs of all $L$ layers in the prefilling stage can then be computed as follows, providing a unified and consistent measure of the overall inference cost across different pruning methods:
\begin{equation}
\sum_{i=1}^{T} \underbrace{(4n_id^2 + 2n_i^2d + 2n_idm)}_{\text{Prefilling FLOPs in layer i}},
\label{method:m_supp1}
\end{equation}

\paragraph{Implementation Details.}
To demonstrate the versatility of HieraVid, we conduct comprehensive evaluations on two representative VideoLLMs, \textit{i.e.}, LLaVA-Video-7B~\cite{zhang2410video} and LLaVA-OneVision-7B~\cite{li2024llava}, and further extend our analysis to Qwen2-VL~\cite{wang2024qwen2}.
Considering that they both have 28 LLM layers, we divide the LLM into $n=4$ stages.
Segment-level merging in Section~\ref{segment-level} is applied before LLMs, followed by a relevance-based pruning stage defined by Equation~\ref{method:m7}, and layer-level balancing~\ref{layer-level} is applied at last two stages.
%
The threshold $\beta$ is set to 0.4, and $\lambda$ is 0.5. All experiments are conducted on NVIDIA A800-PCIe-80GB GPUs.

\subsection{Main Results}

\subsubsection{Comparisons with State-of-the-Art}
Table~\ref{tab:video_models} and Table~\ref{tab:video_models_ov} present the comparison with existing five strong baselines on LLaVA-Video and LLaVA-OneVision under three relative token budgets (30\%, 20\%, and 10\%) to comprehensively assess the performance and robustness of \ourmodel~ under varying pruning intensity.
The experimental results reveal three key strengths of our \ourmodel:

\paragraph{(i) Superior performance:}
Compared with baselins, \ourmodel~demonstrates significant performance improvements across various video understanding benchmarks with different lengths and scene complexities.
As shown in Table~\ref{tab:video_models}, experimental results show that \ourmodel~ outperforms all baseline methods across all benchmarks.
Specifically, at a 30\% retention ratio, \ourmodel~achieves an average accuracy of \textbf{98.4\%}, surpassing the best baseline by \textbf{1.6\%}.
Furthermore, to examine cross-model robustness, we extend \ourmodel~to LLaVA-OneVision. As shown in Table ~\ref{tab:video_models_ov}, \ourmodel~continues to outperform all baselines, achieving \textbf{99.3\%} accuracy at a 30\% retention ratio and maintaining superior performance across all benchmarks.

\paragraph{(ii) Robustness under extreme retention ratios:}
As the retention ratio decreases, all baseline methods observe a notable performance drop. When the token budget is 20\%, our method surpasses FrameFusion by\textbf{ 2.2\%} on LLaVA-Video. Even at a 10\% relevant token budget, \ourmodel~achieves \textbf{92.4\%}, and is the only method remaining above \textbf{90\%}. Similarly, on LLaVA-OneVision, when the retention ratio decreases to 20\%, our method retains \textbf{97.8\%} of the original performance. At a 10\% token budget, it reaches \textbf{94.8\%}, outperforming FastV~\cite{fastv} by \textbf{7.8\%} and exceeding the best baseline by \textbf{2.1\%}.

\paragraph{(iii) Excellence on long video benchmarks:}
Notably, our method effectively handles long videos, as \ourmodel~is designed to efficiently divide videos into segments as described in Section~\ref{segment-level}, preserving both global semantics and local context.
On long-video benchmarks, \textit{e.g.}, VideoMME~\cite{fu2025video} (1-60 mins), as shown in Table~\ref{tab:video_models_ov}, with a 10\% relative token budget,~\ourmodel~surpasses FrameFusion~\cite{fu2024framefusion} up to \textbf{2.6 }points on average, demonstrating the advantage of our method.

\subsubsection{Comparison on Qwen2-VL}
Qwen2-VL~\cite{wang2024qwen2} employs a Naive Dynamic Resolution strategy to flexibly transform frames of varying resolutions into visual tokens. By integrating Multimodal Rotary Position Embedding within a unified framework for both images and videos, it effectively processes long videos exceeding 20 minutes for high-quality question answering, dialogue, and content generation.
We apply \ourmodel{} and several representative pruning methods to Qwen2-VL-7B, and conduct quantitative evaluation on VideoMME. As shown in Table~\ref{supp:qwen2vl}, our method achieves an average accuracy of \textbf{98.0\%} with a 30\% relevant token budget, outperforming FrameFusion by 1.0\%. This result demonstrates that \ourmodel{} can more effectively preserve critical spatiotemporal information while removing redundant tokens, leading to a better trade-off between computational efficiency and model performance.
\subsubsection{Abaltion Study}
To thoroughly evaluate our method, we first validate the proposed components of our \ourmodel~by removing each module sequentially. Then, we compare each component with its own variant in detail. LLaVA-Video with a 30\% token retention ratio is adopted as the default experimental setting.
\begin{table}[t]
\centering
\small
\setlength{\tabcolsep}{6pt}
\renewcommand{\arraystretch}{1.15}
\resizebox{\linewidth}{!}{%
\begin{tabular}{c|ccc|ccc@{}}
\toprule
\multirow{2}{*}{\textbf{Method}} &
\multirow{2}{*}{\textbf{EgoSchema}} &
\multicolumn{2}{c}{\textbf{VideoMME}} &
\multicolumn{2}{c}{\textbf{Avg.}} & \\

\cmidrule(lr){3-4} \cmidrule(lr){5-6} &  & \textbf{w/o sub.} & \textbf{w/ sub.} & \textbf{score} & \textbf{\%} \\
\midrule
Vanilla	
&59.4	&64.1	&71.4	&65.0	&100.0 \\

\midrule
No segment	
&56.2	&60.6	&70.1	&62.3	&95.9 \\
Avg-pool
&56.6	&61.0	&70.4	&62.7	&96.5 \\
Merge Ratio-based segmentation*	
&\textbf{59.2}	&\textbf{62.3}	&\textbf{70.8}	&\textbf{64.1}	&\textbf{98.6} \\

\bottomrule
\end{tabular}
}
\caption{\textbf{Ablation on segment strategy.} Vanilla denotes the original LLaVA-Video before pruning. * means our HieraVid.}
\label{ablation:segment strategy}
\end{table}

\begin{figure}[t]
  \centering
  \includegraphics[width=0.98\linewidth]{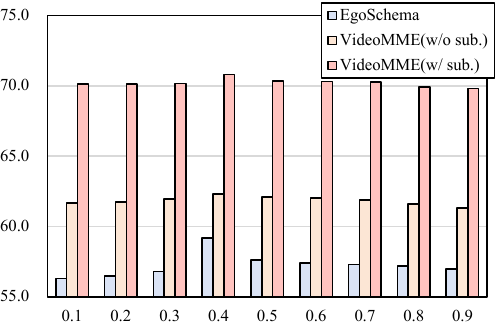} 
  \caption{\textbf{Ablation Experiments} on segment threshold $\beta$.}
  \label{ablation:beta}
\end{figure}

\begin{table}[ht]
\centering
\small
\setlength{\tabcolsep}{6pt}
\renewcommand{\arraystretch}{1.15}
\resizebox{\linewidth}{!}{%
\begin{tabular}{c|ccc|ccc@{}}
\toprule
\multirow{2}{*}{\textbf{Method}} &
\multirow{2}{*}{\textbf{EgoSchema}} &
\multicolumn{2}{c}{\textbf{VideoMME}} &
\multicolumn{2}{c}{\textbf{Avg.}} & \\

\cmidrule(lr){3-4} \cmidrule(lr){5-6} &  & \textbf{w/o sub.} & \textbf{w/ sub.} & \textbf{score} & \textbf{\%} \\
\midrule
Vanilla 
&59.4 &64.1 &71.4 &65.0 &100.0 \\

\midrule
0 &57.5 &61.9 &68.9 &62.8 &96.6 \\
0.2 &58.0 &62.1 &70.1 &63.4 &97.6 \\
0.4 &58.4 &62.2 &\textbf{70.8} &63.8 &98.2 \\
0.5 &\textbf{59.2} &\textbf{62.3} &70.8 &\textbf{64.1} &\textbf{98.6} \\
0.6 &58.0 &62.1 &70.2 &63.4 &97.6 \\
0.8 &57.5 &61.8 &69.6 &63.0 &96.9 \\
1 &57.0 &61.6 &68.5 &62.4 &95.9 \\
\bottomrule
\end{tabular}
}
\caption{\textbf{Ablation study on $\lambda$ for the budget of segment dynamic ratio allocation} on LLaVA-Video (Relative Token Budget: 30\%). Vanilla denotes the original LLaVA-Video before pruning.}
\label{ablation:lambda}
\end{table}

\begin{table}[ht]
\centering
\small
\setlength{\tabcolsep}{6pt}
\renewcommand{\arraystretch}{1.15}
\resizebox{\linewidth}{!}{%
\begin{tabular}{c|ccc|ccc@{}}
\toprule
\multirow{2}{*}{\textbf{Method}} &
\multirow{2}{*}{\textbf{MVBench}} &
\multicolumn{2}{c}{\textbf{NExT-QA}} &

\multicolumn{2}{c}{\textbf{Avg.}} & \\
\cmidrule(lr){3-4} \cmidrule(lr){5-6} &  & \textbf{mc} & \textbf{oe\_test} & \textbf{score} & \textbf{\%} \\
\midrule
Vanilla & 60.4 & 80.2 & 34.6 & 58.4 & 100.0 \\

\midrule
Average Instruction Tokens
& 57.4 & 79.0 & 33.2 & 56.5 & 96.8 \\
Last Instruction Token 
& \textbf{58.3} & \textbf{79.6} & \textbf{33.7} & \textbf{57.3} & \textbf{98.2} \\
\bottomrule
\end{tabular}
}
\caption{\textbf{Ablation on the relevance score}. We compare using the last token against the conventional approach that averages the attention of instruction tokens to visual tokens. Vanilla denotes the original LLaVA-Video before pruning.}
\label{ablation:relevance score}
\end{table}

\paragraph{Ablation on components of~\ourmodel.}
%
%
Table~\ref{ablation:components} presents a detailed analysis of the impact of each proposed module.
First, we remove the layer-level balancing and apply our algorithm only at the LLM input layer. This led to a noticeable overall declines of more than \textbf{4\%}, indicating that layer-level balancing aligns well with the multi-modal information propagation pattern embedded in LLMs.
Next, we remove the segment-level merging, which causes a sharp performance drop on VideoMME (\textbf{1.1} on “w/o sub.” and \textbf{2.3 }on “w/ sub.”). This demonstrates that segment-level merging preserves global semantics of long videos while maintaining local contexts and dynamic characteristics within each segment. 
%
Finally, we replace our frame-level DPP pruning with the token pruning strategy from FastV~\cite{fastv} to evaluate the effectiveness of DPP pruning strategy in \ourmodel. The performance degradation was substantial, \textit{i.e.}, the average performance drops by \textbf{14.6\%} across all benchmarks, indicating that our frame-level DPP pruning avoids visual loss.

\paragraph{Ablation on strategies of segmentation.}
Previous methods~\cite{prunevid,mmg-vid,holitom} typically compute an average pooled feature for each frame, then perform segmentation based on similarities between these frame features. Specifically, PruneVid~\cite{prunevid} employs the density peaks clustering with k-nearest neighbors algorithm to cluster frames and group into temporal segments. MMG-Vid~\cite{mmg-vid} directly computes the similarity between adjacent frame features and defines boundaries using a threshold.
As shown in Table ~\ref{ablation:segment strategy}, compared to the average pooling-based video segmentation, our merge ratio-guided segmentation achieves a \textbf{2.6}-point improvement on the short-video benchmark EgoSchema and a \textbf{1.7}-point improvement in the overall score on the long-video benchmark VideoMME. 
Figure~\ref{fig:visualization_segment} illustrates how the merge ratio quantifies the density between frames and captures inter-segment differences, and highlights the results after segmentation.
In terms of the segmentation threshold $\beta$, a smaller $\beta$ enforces fewer segments and clearer boundaries between them. As shown in Figure~\ref{ablation:beta}, when $\beta$ is set to 0.4, the overall performance on EgoSchema and VideoMME achieves its best balance between inter-segment clarity and intra-segment richness.

\paragraph{Ablation on segment budget allocation.}
For each segment, it is compared against previous and subsequent segments to ensure diversity and representativeness. As shown in Table~\ref{ablation:lambda}, the diversity-only setting ($\lambda=0$) surpasses the representativeness-only setting ($\lambda=1$) by 0.7\%. However, the best results are achieved with a balanced configuration, where $\lambda=0.5$ yields peak scores on both EgoSchema and VideoMME. This confirms that harmonizing diversity and representativeness is vital for VideoLLMs.

\paragraph{Ablation on relevance score.}
In Equation~\ref{method:m7}, the relevance score can be computed using either only the last instruction token or all instruction tokens within the input sequence. Previous studies~\cite{fu2024framefusion,fastv,sparsevlm} adopt the latter approach. Here we compare them in Table~\ref{ablation:relevance score}. When using only the last instruction token, we observe a \textbf{1.4}\% performance improvement. This aligns with the findings of~\cite{zhang2025cross}, which suggest that visual information in MLLMs tends to integrate toward later instruction tokens as layers deepen.

\section{Conclusion}
\label{sec:conclusion}

In this paper, we introduce \ourmodel, a hierarchical video token pruning framework that performs video token pruning from three perspectives: segment-level, frame-level, and layer-level. At the segment-level merging, we propose merge ratio-guided segmentation to ensure intra-segment similarity and inter-segment distinctiveness.
At the layer level, considering the progressive information integration within LLMs, we divide the model into multiple stages to balance information loss from shallow-layer pruning and inefficiency from deep-layer pruning.
For each segment, we propose segment dynamic ratio allocation to dynamically assign pruning budgets.
Finally, we enhance the DPP-based diversity modeling algorithm to balance visual diversity and instruction relevance.
Extensive experiments across multiple VideoLLMs and benchmarks demonstrate that \ourmodel~significantly reduces inference latency while maintaining practically superior performance.















\vfill\eject
\bibliographystyle{ACM-Reference-Format}
\bibliography{sample-base}

\end{document}